\documentclass[10pt,twocolumn,letterpaper]{article}
\usepackage{bm}
\usepackage{cvpr}      % To produce the REVIEW version
\definecolor{cvprblue}{rgb}{0.21,0.49,0.74}
\usepackage[pagebackref,breaklinks,colorlinks,allcolors=cvprblue]{hyperref}

\def\confName{CVPR}
\def\confYear{2027}

\title{HolisticMotion: Holistic Modeling for Motion Reconstruction from Sparse Tracking}

\author{
Runzhen Liu \quad Chuhua Xian \quad Hongmin Cai\\
South China University of Technology\\
Guangzhou, China\\
{\tt\small csalunaticat@mail.scut.edu.cn}\\
{\tt\small chhxian@scut.edu.cn}\\
{\tt\small hmcai@scut.edu.cn}
\and
Fa-Ting Hong\\
The Hong Kong University of Science and Technology\\
Hong Kong SAR\\
{\tt\small fhongac@connect.ust.hk}
}
\begin{document}
\maketitle
\begin{abstract}
Reconstructing full-body motion from sparse tracking signals is crucial for immersive AR/VR applications. Existing methods predominantly adopt a sensor-centric observation perspective, learning full-body representations primarily under sparse conditioning, while additional joint observations available during training remain underutilized as inputs. We propose \textbf{HolisticMotion}, which shifts the modeling perspective toward holistic full-body modeling by treating every joint as a potential input source within a unified skeletal layout. Guided by this holistic perspective, we introduce three complementary designs. First, a curriculum masking strategy bridges full-body motion learning and sparse inference by gradually shifting training from a mixture of masked full-body and sparse inputs to exclusively sparse inputs. Second, body partitioning models region dynamics and helps preserve local motion cues. Third, Kinematic Attention coordinates body regions by combining dynamic spatial interactions with learnable structural priors inspired by the kinematic tree. Experiments on AMASS and HPS demonstrate competitive reconstruction performance across various tracking protocols, with the lowest positional errors among the compared methods. The unified representation also accommodates additional wearable sensors, yielding further improvements in reconstruction accuracy and motion smoothness.
\end{abstract}    
\section{Introduction}
\label{sec:intro}

\begin{figure}[t]
    \centering
    \includegraphics[
        width=\columnwidth,
        trim={0 0mm 0 0mm},
        clip
    ]{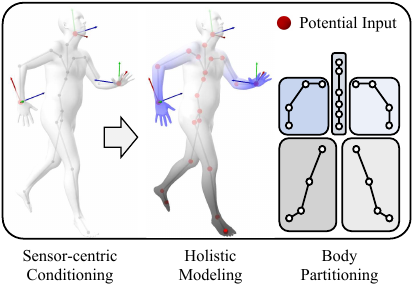}
    \caption{Overview of holistic full-body modeling. Conventional methods reconstruct full-body motion conditioned on sparse sensor observations. HolisticMotion organizes observed and unobserved joints within a unified skeletal layout for full-body modeling. Each joint can provide observations during training, while only tracked joints are observed at inference. Body partitioning further organizes the representation into the torso and four limbs.}
    \label{fig:intro}
    \vspace{-3mm}
\end{figure}

Accurately reconstructing full-body motion from sparse wearable sensors is an ill-posed inverse problem: the model must interpret the observed local motion, infer the unobserved body regions, and coordinate all regions into a coherent full-body action. Existing methods adopt different architectures to address this challenge. AvatarPoser~\cite{jiang2022avatarposer} and EgoPoser~\cite{jiang2024egoposer} directly regress full-body poses from sparse tracking signals. AvatarJLM~\cite{zheng2023realistic} constructs joint-level features from initial full-body pose predictions and models their spatiotemporal interactions, applying random masking to non-anchor tokens during training. SAGE~\cite{feng2024stratified} formulates reconstruction as conditional generation, using HMD signals to guide a diffusion model in generating full-body poses. Despite these advances, existing methods predominantly adopt a sensor-centric observation perspective. Even when explicit joint-level modeling and masking designs are introduced, their observed inputs remain confined to the tracked joints, with ground-truth signals from untracked joints providing supervision rather than additional input observations. Their full-body representations are therefore learned under sparse conditioning. This motivates exploiting additional joint observations available during training as partially visible inputs for learning cross-joint dependencies, before adapting to the sparse inference setting.

To this end, we propose \textbf{HolisticMotion}, which shifts the modeling perspective for sparse-tracking reconstruction from sensor-centric conditioning to holistic full-body modeling by unifying observed and unobserved joints within a single input representation, as illustrated in Fig.~\ref{fig:intro}. We allocate an input position to every skeletal joint, treating each joint as a potential source of observations within a unified representation. Inspired by MAE~\cite{he2022masked}, we use joint-wise masking to specify which signals are visible. This formulation casts sparse-tracking reconstruction as a special case of masked pose reconstruction, where only the tracked joints are observed. The unified representation also accommodates additional wearable trackers by incorporating their signals at the corresponding joint positions. Building on this formulation, we introduce \textbf{Masked Pose Modeling Guidance} (MPMG), a curriculum strategy designed to bridge full-body motion learning and sparse inference. For masked full-body samples, the head and hands remain visible, while randomly retained observations from other joints expose the network to richer motion relationships across the skeleton. By gradually decreasing the proportion of these masked samples in favor of exclusively sparse inputs, MPMG encourages the model to internalize holistic motion dependencies before progressively adapting to the sparse setting encountered at inference.

Within this holistic perspective, full-body modeling must balance global interactions with the preservation of local motion cues. The head and hands provide direct observations, whereas the unobserved regions introduce greater uncertainty. Directly mixing all joint features through global attention may dilute reliable local cues by combining them with uncertain representations. To address this, we draw on body-partitioning strategies from general full-body modeling~\cite{du2015hierarchical, guo2020action2motion}, dividing the body into five functional regions (the torso and four limbs, as shown in Fig.~\ref{fig:intro}). This regional division enables the network to learn region-specific dynamics while mitigating feature interference.

While functional partitioning supports regional feature learning, these regions must still be coordinated to produce a coherent full-body action. In skeleton-based action modeling~\cite{yan2018spatial, shi2019two}, joint graphs commonly encode the human skeleton's kinematic tree to organize information propagation along anatomical connections. Inspired by this organization, we introduce \textbf{Kinematic Attention} to coordinate interactions among functional body regions. This module combines an input-dependent dynamic branch with a learnable structural branch initialized from a torso-centered topology connecting the torso to the four limbs. The resulting structural prior remains adaptable during learning and complements dynamic spatial interactions to promote coherent cross-region coordination.

Our main contributions are:
\begin{itemize}
\item We formulate sparse-tracking reconstruction as masked full-body observation modeling, where every skeletal joint is a potential source of training observations. Within this formulation, \textbf{MPMG} bridges full-body motion learning and sparse inference through curriculum training.

\item We adopt body partitioning to model region-specific dynamics and introduce \textbf{Kinematic Attention} to coordinate interactions across body regions. By combining dynamic spatial interactions with learnable structural priors inspired by the kinematic tree, this design promotes coherent full-body motion.
\item Extensive experiments on AMASS~\cite{mahmood2019amass} and HPS~\cite{guzov2021human} demonstrate competitive reconstruction performance across various tracking protocols, achieving highly accurate spatial reconstruction and yielding the lowest positional errors among the compared methods. Our unified representation also accommodates additional wearable sensors, yielding further improvements in accuracy and motion smoothness.
\end{itemize}
\section{Related Work}
\label{sec:relatedwork}

\begin{figure*}[!t]
    \centering
    \includegraphics[width=1\linewidth,trim={0 0mm 0 0mm},clip]{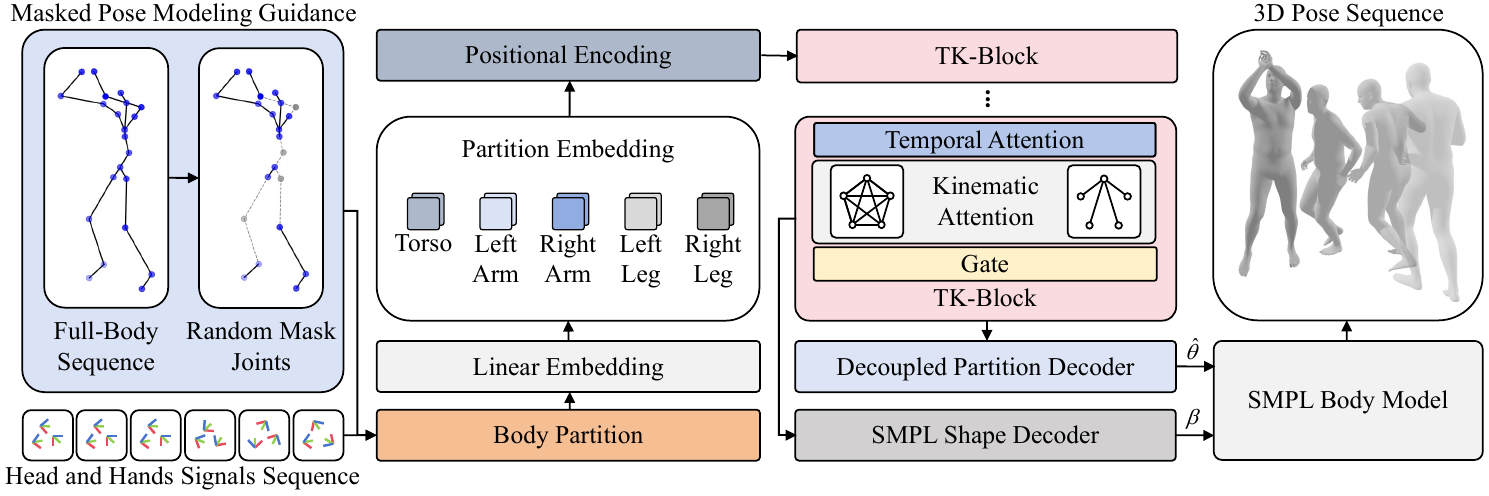}
    \caption{Illustration of our framework. During training, MPMG guides the learning of body-part features from masked full-body and sparse inputs. The partition features are augmented with positional encodings and modulated before entering stacked TK-Blocks, which combine temporal attention with gated kinematic attention to model full-body motion. Separate pose and shape decoders predict SMPL parameters to reconstruct the 3D pose sequence.}
    \vspace{-3mm}
    \label{fig:framework}
\end{figure*}

\noindent\textbf{Full-Body Pose Estimation from Sparse Inputs.} Traditional methods~\cite{goldenberg2003complete,FinalIK,parger2018human} employ inverse kinematics (IK) solvers and heuristic rules to calculate the variable joint parameters to produce a desired end-effector location. Many prior works ~\cite{huang2018deep,jiang2022transformer,von2017sparse,yi2021transpose,yi2022physical} focus on reconstructing full-body motion from six inertial measurement units (IMUs). With the rise of VR/AR applications, and considering that common consumer-grade VR/AR devices, such as head-mounted devices (HMDs), only provide motion information for the user's head and hands, many recent methods ~\cite{ahuja2021coolmoves,yang2021lobstr,dittadi2021full,jiang2022avatarposer,zheng2023realistic,aliakbarian2022flag,barquero2025sparse,dai2024hmd,feng2024stratified,jiang2024egoposer,winkler2022questsim} have shifted focus to this sparser setting. CoolMoves~\cite{ahuja2021coolmoves} first estimates the full-body pose using only tracking signals from the head and hands. LoBSTr~\cite{yang2021lobstr} predicts lower-body poses from upper-body signals. HMD-Poser~\cite{dai2024hmd} supports multiple input configurations, in addition to the standard HMD input. EgoPoser~\cite{jiang2024egoposer} models the cameras' FoV to simulate realistic hand tracking signal patterns. However, these approaches predominantly operate under a sensor-centric paradigm, learning full-body representations strictly under sparse conditioning. To overcome this, our formulation shifts to a holistic perspective, incorporating additional joint observations as partially visible training inputs within a unified skeletal layout before adapting to sparse inference.

\noindent\textbf{Deep Learning for Human Pose Estimation.} Recent research predominantly adopts deep learning approaches. For instance, VAEHMD~\cite{dittadi2021full} proposes a method based on variational autoencoders to generate articulated poses of a human skeleton based on noisy streams of head and hand pose. LoBSTr~\cite{yang2021lobstr} uses GRU-based recurrent architecture to predict the lower-body pose. AvatarPoser~\cite{jiang2022avatarposer} utilizes a Transformer~\cite{vaswani2017attention} encoder to predict full-body poses, complemented by a separate IK module to adjust the arm positions. AvatarJLM~\cite{zheng2023realistic} employs alternating spatial and temporal transformer blocks to model the correlations of joint-level features. SAGE~\cite{feng2024stratified} uses a stratified diffusion model~\cite{ho2020denoising} to first predict the upper-body pose, which then serves as a condition for estimating the lower-body pose. While these methods primarily focus on sparse-input conditioning, our holistic formulation motivates incorporating established designs from full-body motion modeling: body partitioning for regional dynamics and skeletal structural priors for cross-region coordination.

\section{Method}

\subsection{Overview}
We aim to reconstruct full-body motion $\bm{\Theta} \in \mathbb{R}^{T \times J \times S}$ from sparse tracking signals $\bm{X} \in \mathbb{R}^{T \times N \times C}$ from HMDs and hand controllers. We represent the full-body pose using the first $J=22$ joints of the SMPL-H model~\cite{loper2023smpl}, and employ a Forward Kinematics (FK) module to calculate 3D joint positions. Following~\cite{jiang2022avatarposer}, our basic inputs are 6-DoF head and hand signals, corresponding to a per-joint feature dimension of $C = 3+3+6+6 = 18$ (comprising 3D position, 3D linear velocity, 6D rotation, and 6D angular velocity). We adopt this continuous 6D representation for rotations due to its simplicity and continuity for neural network learning~\cite{zhou2019continuity}. In the original task formulation, $N=3$, corresponding to the head and both hands tracked by an HMD and hand controllers. The network directly outputs the 6D rotation representations ($S=6$), which are subsequently converted into 3D axis-angle formats for the FK module. Furthermore, to address the limitation of mean-shape assumptions in prior works~\cite{dittadi2021full, jiang2022avatarposer}, we synthesize tracking observations using real body shapes and additionally predict the 16-dimensional SMPL shape $\beta \in \mathbb{R}^{T \times 16}$ to account for individual body proportions.

\subsection{Input Processing}

\noindent\textbf{Holistic Modeling.}
To establish a unified full-body representation, we expand the sparse tracking signals $\bm{X} \in \mathbb{R}^{T \times N \times C}$ into a complete skeletal layout before feature extraction. Given a skeleton with $J$ joints, we define the observed set $\mathcal{O}$ as the head and hands, with $|\mathcal{O}|=N$, and the unobserved set $\mathcal{U}$ as the remaining joints. We assign the observed features to their corresponding joint indices and zero-pad the unobserved joints, yielding $\bm{X}_{\mathrm{holistic}} \in \mathbb{R}^{T \times J \times C}$. 

\begin{figure*}[t]
    \centering
    \includegraphics[width=\linewidth]{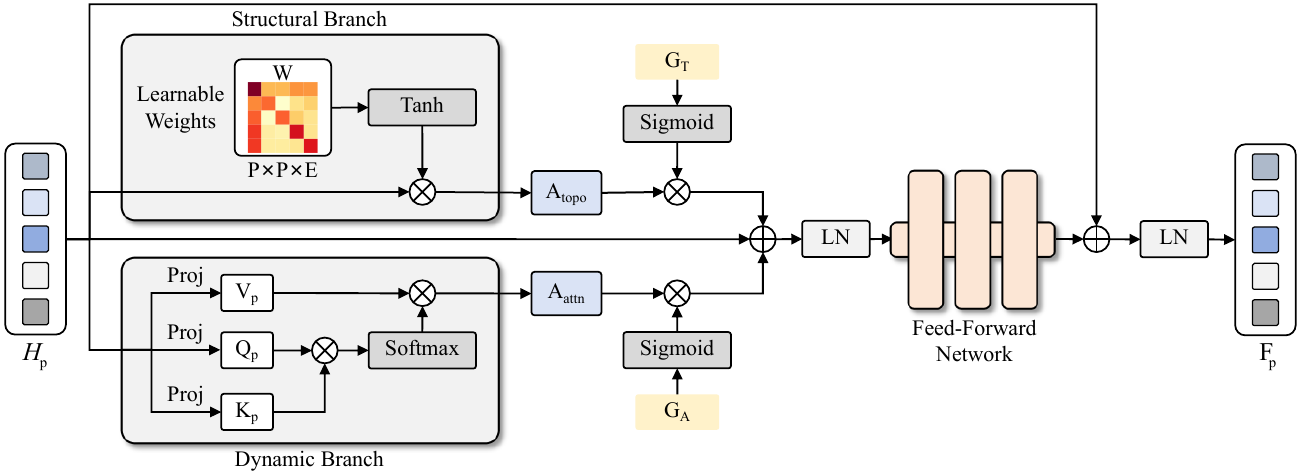}
    \caption{Architecture of Kinematic Attention. Dynamic and structural branches model input-dependent interactions and topology-guided regional relationships, respectively. Their outputs are combined through learnable gates and refined through residual connections, layer normalization, and a feed-forward network.}
    \label{fig:kattn}
    \vspace{-5mm}
\end{figure*}

\noindent\textbf{Masked Pose Modeling Guidance.}
To effectively learn full-body dependencies, inspired by masked modeling~\cite{tong2022videomae,zhu2023motionbert}, we introduce Masked Pose Modeling Guidance (MPMG). Since head and hand trajectories reveal crucial motion intent, we preserve them as conditioning anchors while randomly masking a subset of the remaining joints. Crucially, each selected joint is masked across the entire temporal window. This prevents trivial temporal interpolation, forcing the network to explicitly reconstruct poses by exploiting spatial-temporal cross-joint correlations.

Formally, MPMG mixes masked full-body inputs with sparse head and hand inputs. For the full-body sequence $\bm{X}_{\mathrm{full}}\in\mathbb{R}^{T\times J\times C}$, we preserve the anchor joints ($m_j=1,\forall j\in\mathcal{O}$) and randomly retain the remaining joints ($m_j\sim\operatorname{Bernoulli}(0.5),\forall j\in\mathcal{U}$). The network input $\bm{X}_{\mathrm{in}}\in\mathbb{R}^{T\times J\times C}$ is sampled as:
\begin{equation}
    \bm{X}_{\mathrm{in}} =
    \begin{cases}
        \bm{X}_{\mathrm{full}}\odot\bm{m}, & \text{with probability } q(s),\\
        \bm{X}_{\mathrm{holistic}}, & \text{with probability } 1-q(s),
    \end{cases}
\end{equation}
where $\bm{m}$ is broadcast across the temporal and feature dimensions, and $q(s)$ decreases linearly to zero with training step $s$. This curriculum combines full-body motion learning with sparse-input reconstruction, gradually shifting training to exclusively $\bm{X}_{\mathrm{holistic}}$ to match inference.

\noindent\textbf{Functional Body Partitioning.}
Building upon the full-body representation, we partition the input sequence into five functional regions to organize features for subsequent modeling, as illustrated in Fig.~\ref{fig:intro}. We partition \(X_{\mathrm{in}}\) into the torso and four limbs, indexed by \(p\in\{1,\ldots,5\}\) in the order torso, left arm, right arm, left leg, and right leg. The torso contains six joints, while each limb contains four.

% \begin{figure*}[!t]
%     \centering
%     \includegraphics[width=1\linewidth,trim={0 0mm 0 0mm},clip]{figs/GSMandTK.pdf}
%     \caption{Illustration of Global Synchronized Modulation and Temporal-Kinematic Block.}
%     \vspace{-5mm}
%     \label{fig:GSMandTK}
% \end{figure*}

\subsection{Spatio-Temporal Kinematic Modeling}

\noindent\textbf{Feature Initialization.}
We first embed each body partition and add positional encodings:
\begin{equation}
    \bm{E}_p = f_p^{\mathrm{embed}}(\bm{X}_p) + \mathrm{PE}.
\end{equation}
An MLP takes the concatenated partition features and predicts region-specific scale and shift parameters:
\begin{equation}
    [\alpha_1, \beta_1, \dots, \alpha_5, \beta_5]
    = f^{\mathrm{mod}}(\mathrm{Concat}(\bm{E}_1, \dots, \bm{E}_5)).
\end{equation}
These parameters modulate each partition before spatio-temporal modeling:
\begin{equation}
    \bm{E}'_p = \bm{E}_p \odot (1 + \alpha_p) + \beta_p,
    \quad p \in \{1, \dots, 5\},
\end{equation}
where $\odot$ denotes element-wise multiplication.

Next, the modulated features $\bm{E}'_p$ are processed by the \textbf{Temporal-Kinematic Blocks (TK-Blocks)}, which serve as the primary computational unit for motion synthesis. The TK-Blocks interleave temporal intent modeling with spatial kinematic interactions to reconstruct accurate motions.

\noindent\textbf{Temporal Attention with Symmetric Sharing.}
We apply temporal self-attention separately to the five partitions to model their respective motion dynamics. To enhance optimization efficiency and capture the isomorphic temporal patterns inherent in human locomotion, symmetric limb partitions share identical parameters. Let $\bm{F}^i_p$ denote the input features for partition $p$ at layer $i$, where $\bm{F}^1_p = \bm{E}'_p$. The temporal attention is defined as:
\begin{equation}
    \bm{H}^i_p = \text{T-MSA}_{\theta^i_{\lfloor p/2 \rfloor + 1}}(\bm{F}^i_p), \quad p \in \{1, \dots, 5\},
\end{equation}
where $\text{T-MSA}$ identifies the temporal multi-head self-attention. Through the index $\lfloor p/2 \rfloor + 1$, the torso ($p=1$) utilizes unique parameters $\theta_1$, while the paired arms ($p \in \{2, 3\}$) and legs ($p \in \{4, 5\}$) share parameters $\theta_2$ and $\theta_3$ respectively.

\noindent\textbf{Kinematic Attention.}
While data-driven spatial attention effectively captures dynamic motion correlations~\cite{zheng2023realistic,zhu2023motionbert}, standard self-attention lacks explicit structural inductive biases. Relying solely on unconstrained data-driven interactions treats all body regions symmetrically, essentially overlooking the inherent hierarchical dependencies defined by the human skeleton. To address this, we introduce a dual-branch Kinematic Attention mechanism that synergizes input-dependent dynamic interactions with a learnable structural prior, as illustrated in Fig.~\ref{fig:kattn}.

\noindent\textit{(i) Dynamic Branch.}
This branch extracts synergistic correlations during complex movements using spatial multi-head self-attention (S-MSA) across the partition dimension:
\begin{equation}
    \bm{A}^i_{\mathrm{attn}} = \text{S-MSA}([\bm{H}^i_1, \dots, \bm{H}^i_5]).
\end{equation}

\noindent\textit{(ii) Structural Branch.}
Inspired by the skeleton's kinematic tree, this branch uses a learnable tensor
$\bm{W}^i_{\mathrm{topo}} \in \mathbb{R}^{P \times P \times E}$,
initialized as a star-shaped kinematic tree rooted at the torso. A tanh activation regulates the influence of these kinematic correlations:
\begin{equation}
    \bm{A}^i_{\mathrm{topo}}
    = \tanh(\bm{W}^i_{\mathrm{topo}})
    \otimes \bm{H}^i,
\end{equation}
where $\otimes$ denotes channel-wise weighted aggregation across body regions.

\noindent\textbf{Gated Fusion and Feature Update.}
To balance the contributions of dynamic interactions and structural aggregation across body regions and network depths, each TK-Block learns region-specific gating vectors $\bm{G}_A^i$ and $\bm{G}_T^i$. The fused features are then processed through residual connections, Layer Normalization (LN), and a Feed-Forward Network (FFN):
\begin{equation}
\begin{aligned}
    \bm{Z}^i &= \text{LN}\big(\bm{H}^i + \sigma(\bm{G}^i_A) \bm{A}^i_{\text{attn}} + \sigma(\bm{G}^i_T) \bm{A}^i_{\text{topo}}\big), \\
    \bm{F}^{i+1}_p &= \text{LN}\big(\bm{Z}^i_p + \text{FFN}^i(\bm{Z}^i_p)\big).
\end{aligned}
\end{equation}

\subsection{SMPL Pose and Shape Estimation}
\noindent\textbf{Decoupled Partition Decoder.} Prior methods~\cite{jiang2022avatarposer,jiang2024egoposer,dai2024hmd} typically map backbone features to final poses using centralized architectures, which often suffer from feature entanglement and gradient interference. To preserve semantic clarity, we instead assign a dedicated MLP head to each of the five partitions. Consistent with our temporal attention design, we share decoder weights for paired limbs. Given the output features $\mathbf{F}^L = \{\mathbf{f}_p^L\}_{p=1}^5$ from the final TK-Block, the decoding is formulated as:
\begin{equation}
\begin{aligned}
    \hat{\mathbf{Y}}_p = D_{\phi_{\lfloor p/2 \rfloor + 1}}(\mathbf{f}_p^L), \quad p \in \{1, 2, 3, 4, 5\},
\end{aligned}
\end{equation}
where $D_{\phi}$ denotes the decoding head parameterized by $\phi$. The final pose $\hat{\mathbf{Y}}$ is obtained by concatenating all local predictions. As the SMPL model input, 6D $\hat{\mathbf{Y}}$ are then converted into 3D axis-angle representations, denoted as $\hat{\mathbf{\theta}}$.

\noindent\textbf{SMPL Shape Estimation.} Following HMD-Poser~\cite{dai2024hmd}, we estimate the SMPL shape parameters \(\beta\) using an MLP that takes the concatenated final features from all five body partitions. The predicted shape \(\beta\) and pose \(\hat{\theta}\) are passed to SMPL to obtain joint positions, through which shape is optimized without direct \(\beta\) supervision.

\subsection{Loss Function}

We supervise root orientation, local joint rotations, joint positions, velocities, and accelerations using L1 losses, denoted by \(\mathcal{L}_{\mathrm{ori}}\), \(\mathcal{L}_{\mathrm{rot}}\), \(\mathcal{L}_{\mathrm{pos}}\), \(\mathcal{L}_{\mathrm{vel}}\), and \(\mathcal{L}_{\mathrm{acc}}\), respectively. Shape predictions are constrained by an L2 regularization loss \(\mathcal{L}_{\mathrm{reg}}\). To encourage temporal consistency, we penalize deviations of per-frame shape predictions from their window mean:
\begin{equation}
\begin{aligned}
    \mathcal{L}_{consist} = \frac{1}{T} \sum_{t=1}^{T} \| \beta_t - \bar{\beta} \|_1.
\end{aligned}
\end{equation}

\section{Experiments}
\label{sec:exp}

\begin{table*}[!t]
    \centering
    \caption{Quantitative comparison with state-of-the-art on AMASS in the standard mode.}
    \setlength{\tabcolsep}{4pt}
    \begin{tabular}{lcccc|cccc}
    \toprule
                 & \multicolumn{4}{c}{AMASS-P1} & \multicolumn{4}{c}{AMASS-P2} \\
         Methods & MPJRE $\downarrow$ & MPJPE $\downarrow$ & MPJVE $\downarrow$ & Jitter $\downarrow$ & MPJRE $\downarrow$ & MPJPE $\downarrow$ & MPJVE $\downarrow$ & Jitter $\downarrow$ \\
    \midrule
    
         AvatarJLM & 3.01 & 3.35 & 20.96 & 10.08 & 5.21 & \underline{5.23} & \textbf{22.72} & \underline{1.77} \\
         SAGE & 2.49 & 3.18 & 20.78 & 6.46 & 4.72 & 5.92 & 33.29 & 8.16 \\
         EgoPoser & 2.90 & 3.80 & 24.26 & 14.89 & 4.84 & 6.26 & 33.96 & 12.29 \\
         HMD-Poser & 2.29 & 3.15 & 17.47 & 6.43 & 4.93 & 6.06 & 31.21 & 7.30 \\
         RPM & 3.25 & 4.08 & 19.21 & \textbf{4.21} & 5.15 & 6.56 & 27.16 & \textbf{1.55} \\
    \midrule

        HolisticMotion-40 & \textbf{2.20} & \underline{3.00} & \underline{16.08} & 6.63 & \underline{4.67} & 5.50 & 30.25 & 7.61 \\
        HolisticMotion-80 & \underline{2.22} & \textbf{2.95} & \textbf{14.88} & \underline{6.02} & \textbf{4.66} & \textbf{5.08} & \underline{27.16} & 6.53 \\
    \bottomrule
    \end{tabular}
    \label{tab:comparison_standard}
\end{table*}

\begin{figure*}[t]
    \centering
    \includegraphics[width=1\linewidth,trim={0 0mm 0 0mm},clip]{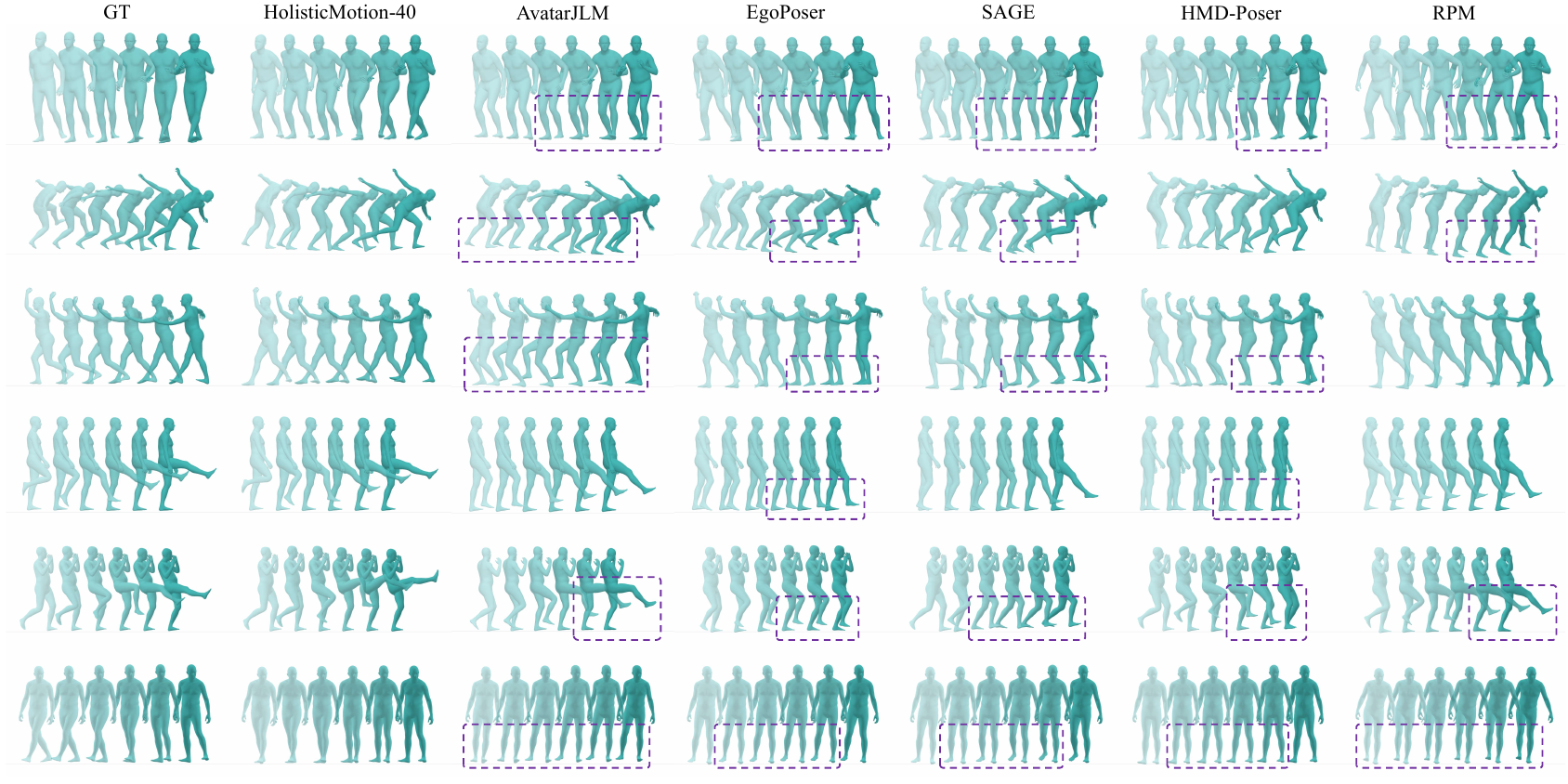}
    \caption{Visual comparisons on the AMASS-P1 in the standard mode.}
    \vspace{-5mm}
    \label{fig:comparison}
\end{figure*}

% In this section, we first introduce the implementation details of our experiments. Then, we compare our approach with state-of-the-art methods and conduct ablation studies on the public AMASS~\cite{mahmood2019amass} dataset. Finally, we conduct further analysis and discussion of our method through additional experiments.
\subsection{Experimental Setup}
\noindent\textbf{Implementation details.} To comprehensively evaluate our method, we adopt two input window size settings: $t = 40$, following HMD-Poser~\cite{dai2024hmd}, and $t = 80$, following EgoPoser~\cite{jiang2024egoposer}. For both settings, we employ a sliding window strategy during inference, processing head and hand signals across the entire window to predict the full-body pose for the current frame. Furthermore, we simulate a hand tracking mode from EgoPoser~\cite{jiang2024egoposer}. 

\noindent\textbf{Metrics.} Similarly to previous works~\cite{jiang2022avatarposer,dai2024hmd,feng2024stratified}, we use the MPJRE (Mean Per Joint Rotation Error, in degrees), the MPJPE (Mean Per Joint Position Error, in $cm$), and the MPJVE (Mean Per Joint Velocity Error, in $cm/s$). To measure the smoothness of the motion, we compute the Jitter ($10^2m/s^3$) as in~\cite{du2023avatars}.

\begin{table}[t]
    \centering
    \caption{Inference speed and quantitative comparison on HPS.}
    \label{tab:comparison_hps}
    \small
    \setlength{\tabcolsep}{3pt}
    \begin{tabular}{lccccc}
        \toprule
        Method & MPJRE$\downarrow$ & MPJPE$\downarrow$ & MPJVE$\downarrow$ & Jitter$\downarrow$ & FPS$\uparrow$ \\
        \midrule
        AvatarJLM      & 10.17 & 20.71 & 35.59 & 6.08  & 17.3 \\
        EgoPoser       & 7.41  & 6.93  & 28.88 & 5.55  & \textbf{101.2} \\
        HMD-Poser      & 9.25  & 27.74 & 76.61 & 12.32 & 44.6 \\
        RPM            & 9.18  & 16.44 & 53.93 & 18.46 & 99.9 \\
        SAGE           & 9.47  & 12.76 & 34.77 & 14.46 & 15.0 \\
        \midrule
        HolisticMotion-80 & \textbf{7.29} & \textbf{6.57} & \textbf{25.91} & \textbf{4.79} & 84.5 \\
        \bottomrule
    \end{tabular}
\end{table}

\noindent\textbf{Datasets.} Following common practice~\cite{aliakbarian2022flag,jiang2022avatarposer,dai2024hmd,barquero2025sparse}, we evaluate our method on the AMASS~\cite{mahmood2019amass} dataset using two standard protocols. Protocol 1 utilizes the CMU~\cite{AMASS_CMU}, BMLr~\cite{troje2002decomposing}, and HDM05~\cite{muller2007mocap} subsets, randomly split into 90 percent training and 10 percent testing data. Protocol 2 adopts a more extensive setting, employing twelve subsets~\cite{AMASS_ACCAD,ghorbani2020movi,troje2002decomposing,AMASS_CMU,mandery2015kit,mandery2016unifying,krebs2021kit,AMASS_EyesJapanDataset,muller2007mocap,AMASS_MoSh,AMASS_PosePrior,AMASS_SFU,AMASS_TotalCapture} for training while using HumanEva~\cite{sigal2010humaneva} and Transition~\cite{mahmood2019amass} for evaluation. To evaluate performance in the wild, we additionally evaluate our method on the HPS~\cite{guzov2021human} dataset, which records human motion in large-scale scenes.

\begin{table*}[t]
    \centering
    \caption{Quantitative comparison with state-of-the-art on AMASS in the hand tracking mode.}
    \setlength{\tabcolsep}{4pt}
    \begin{tabular}{lcccc|cccc}
    \toprule
                 & \multicolumn{4}{c}{AMASS-P1} & \multicolumn{4}{c}{AMASS-P2} \\
         Methods & MPJRE $\downarrow$ & MPJPE $\downarrow$ & MPJVE $\downarrow$ & Jitter $\downarrow$ & MPJRE $\downarrow$ & MPJPE $\downarrow$ & MPJVE $\downarrow$ & Jitter $\downarrow$ \\
    \midrule
         EgoPoser & 4.75 & 6.14 & 33.36 & 17.28 & \textbf{6.40} & \underline{9.36} & 51.76 & 27.02 \\
         RPM & 4.53 & 6.50 & 26.10 & \textbf{4.40} & 7.60 & 11.84 & \textbf{38.93} & \textbf{1.72} \\
    \midrule

        HolisticMotion-40 & \underline{3.29} & \underline{4.44} & \underline{26.08} & 12.38 & 7.07 & 9.49 & 52.90 & 19.90 \\
        HolisticMotion-80 & \textbf{3.22} & \textbf{4.25} & \textbf{22.41} & \underline{9.60} & \underline{6.54} & \textbf{8.56} & \underline{45.04} & \underline{14.97} \\
    \bottomrule
    \end{tabular}
    \label{tab:comparison_ht}
\end{table*}

\begin{figure*}[t]
    \centering
    \includegraphics[width=1\linewidth,trim={0 0mm 0 0mm},clip]{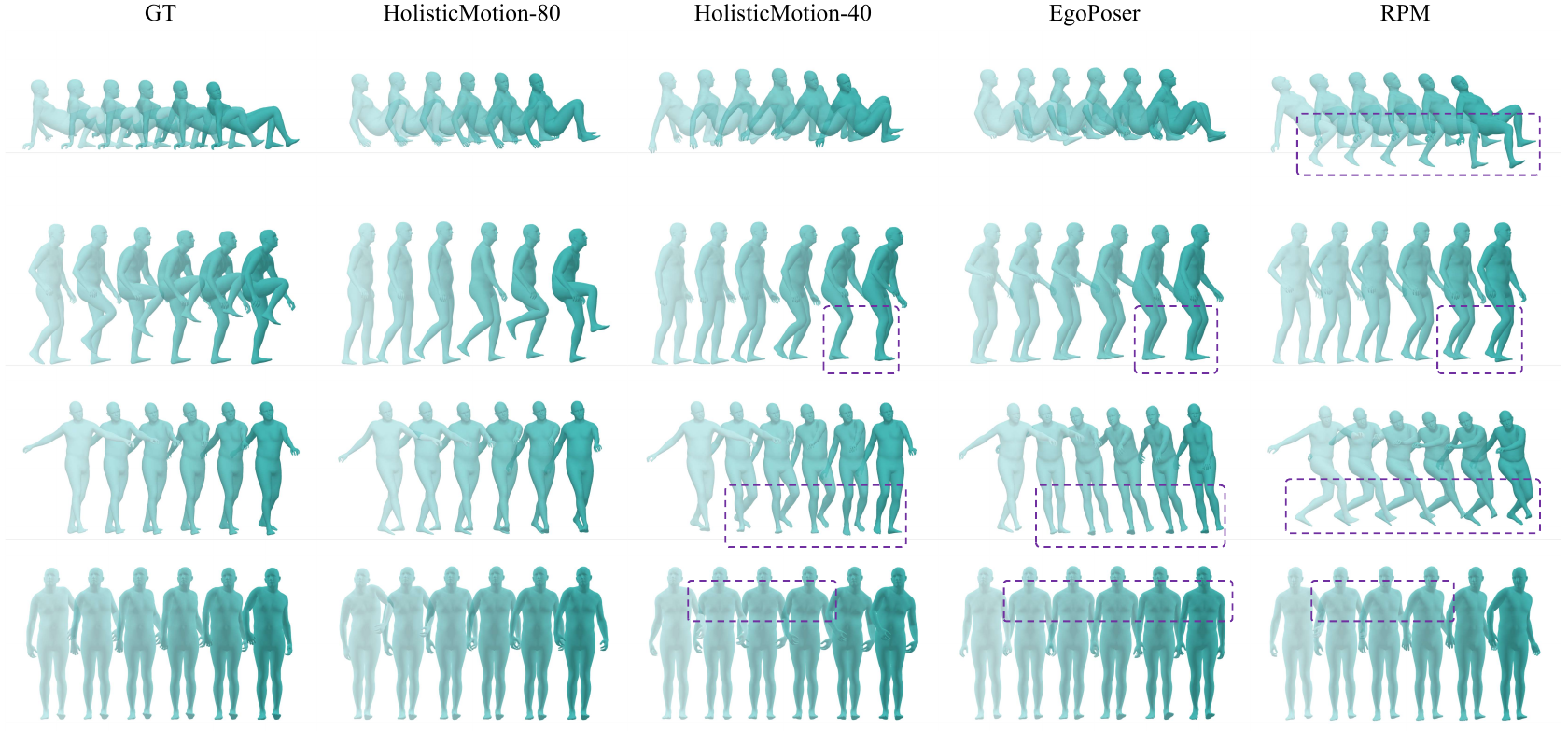}
    \caption{Visual comparisons on the AMASS-P1 in the hand tracking mode.}
    \vspace{-5mm}
    \label{fig:comparison_ht}
\end{figure*}

\subsection{Comparison}

We compare HolisticMotion with AvatarJLM~\cite{zheng2023realistic}, SAGE~\cite{feng2024stratified}, EgoPoser~\cite{jiang2024egoposer}, HMD-Poser~\cite{dai2024hmd}, and RPM~\cite{barquero2025sparse} on AMASS-P1 and AMASS-P2. We evaluate our method using 40- and 80-frame input windows, denoted as HolisticMotion-40 and HolisticMotion-80, respectively. We further compare with EgoPoser and RPM under the hand tracking setting, where hand observations may become unavailable outside the headset's field of view.

\noindent\textbf{Standard Mode.} As shown in Table~\ref{tab:comparison_standard}, HolisticMotion achieves strong spatial reconstruction performance under the standard setting. On AMASS-P1, HolisticMotion-40 achieves the lowest MPJRE, while HolisticMotion-80 obtains the lowest MPJPE and MPJVE. On AMASS-P2, HolisticMotion-80 achieves the lowest MPJRE and MPJPE among the compared methods.

\noindent\textbf{Hand Tracking Mode.} Under the hand tracking setting (Table~\ref{tab:comparison_ht}), HolisticMotion-80 achieves the lowest MPJRE, MPJPE, and MPJVE on AMASS-P1 and the lowest MPJPE on AMASS-P2. Increasing the input window from 40 to 80 frames reduces all four errors on both protocols, suggesting that longer temporal context helps compensate for intermittent hand observations. RPM uses explicit temporal smoothing to improve motion continuity, but exhibits higher rotational and positional errors than HolisticMotion-80, reflecting a trade-off between smoothness and spatial accuracy. Visual comparisons in Figs.~\ref{fig:comparison} and~\ref{fig:comparison_ht} illustrate the reconstructed motion sequences, with colors progressing from light to dark over time.

\noindent\textbf{Large-Scale Scenes.} Following EgoPoser~\cite{jiang2024egoposer}, we evaluate cross-dataset generalization on HPS using models trained only on AMASS-P1. As shown in Table~\ref{tab:comparison_hps}, HolisticMotion-80 achieves the lowest errors across all four reconstruction metrics among the compared methods. These results demonstrate that its spatial reconstruction advantages extend to HPS while also improving temporal accuracy and motion smoothness.

\noindent\textbf{Inference Efficiency.} We further compare inference speed on an NVIDIA RTX 5060 Ti. AvatarJLM performs computationally expensive spatial attention over joint-level tokens, whereas HolisticMotion models spatial interactions among five body regions, reducing the token count for spatial attention. HolisticMotion-80 runs at 84.5 FPS, compared with 17.3 FPS for AvatarJLM and 15.0 FPS for SAGE. Although EgoPoser and RPM achieve higher inference speeds of 101.2 and 99.9 FPS, respectively, HolisticMotion-80 obtains lower errors across all four reconstruction metrics on HPS, offering a favorable balance between reconstruction quality and inference efficiency.

\subsection{Ablation Studies}
\begin{figure*}[t]
    \centering
    \includegraphics[width=1\linewidth,trim={0 0mm 0 0mm},clip]{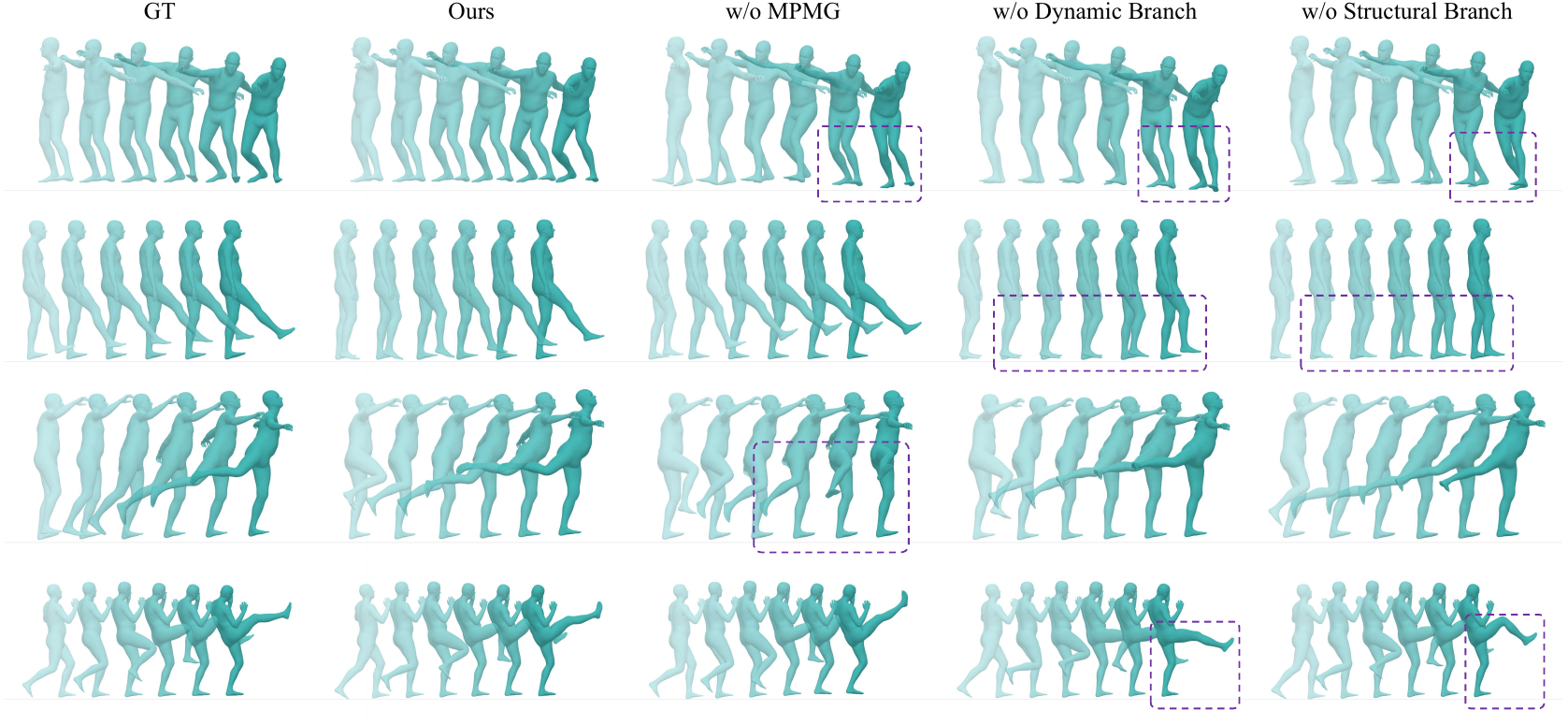}
    \caption{Visual comparison of ablation studies on the AMASS-P1.}
    \label{fig:ablation}
    \vspace{-3mm}
\end{figure*}

We perform ablation experiments on AMASS-P1 to evaluate the effectiveness of each design. The results are shown in Table~\ref{tab:ablation_study} and visual comparisons are shown in Fig.~\ref{fig:ablation}.

\begin{table}[t]
    \begin{minipage}[t]{0.48\textwidth}
        \centering   
        \caption{Ablation results on the AMASS-P1.}
        \label{tab:ablation_study}
        \setlength{\tabcolsep}{3pt}
        \resizebox{\textwidth}{!}{
        \begin{tabular}{l|cccc}
        \toprule
        Configurations & MPJRE $\downarrow$ & MPJPE $\downarrow$ & MPJVE $\downarrow$ & Jitter $\downarrow$ \\
        \midrule 
        w/o MPMG & 2.25 & 3.01 & 15.04 & 5.94 \\
        w/o Dynamic Branch & 2.24 & 2.99 & \textbf{14.86} & \textbf{5.76} \\
        w/o Structural Branch & 2.22 & 2.96 & 15.04 & 6.13\\
        w/o Decoupled Decoder & 2.22 & 2.95 & 15.43 & 6.38 \\
        \midrule
        Ours  & \textbf{2.22} & \textbf{2.95} & 14.88 & 6.02 \\
        \bottomrule
        \end{tabular}
        }
    \end{minipage}
\end{table}

\noindent\textbf{Masked Pose Modeling Guidance.}
Training exclusively with sparse head and hand inputs without MPMG (Row 1) produces the largest increase in MPJRE and MPJPE among the component ablations. MPJVE also increases, whereas Jitter slightly decreases. These results support the benefit of learning from richer body observations before adapting to sparse inputs, with a minor trade-off in motion smoothness.

\noindent\textbf{Decoupled Decoder.}
Replacing the decoupled decoder with a centralized head (Row 4) increases MPJVE and Jitter. Specifically, the decoupled decoder reduces these metrics by 3.6\% and 5.6\%, respectively, indicating improved temporal accuracy and smoothness while leaving spatial errors nearly unchanged.

\begin{table}[t]
    \begin{minipage}[t]{0.48\textwidth}
    \centering
    \caption{Ablation of partitioning strategies on the AMASS-P1.}
    \label{tab:ablation_partition}
    \setlength{\tabcolsep}{3pt}
    \resizebox{\textwidth}{!}{
    \begin{tabular}{lcccc}
        \toprule
        Partition & MPJRE $\downarrow$ & MPJPE $\downarrow$ & MPJVE $\downarrow$ & Jitter$\downarrow$ \\
        \midrule
        Single body       & 2.58 & 3.53 & 17.89 & 7.22 \\
        Upper/lower       & 2.37 & 3.18 & 15.22 & \textbf{5.58} \\
        Left/middle/right & 2.36 & 3.26 & 15.58 & 6.19 \\
        Shuffled          & 2.32 & 3.13 & 14.96 & 5.60 \\
        \midrule
        Ours              & \textbf{2.22} & \textbf{2.95} & \textbf{14.88} & 6.02 \\
        \bottomrule
    \end{tabular}
    }
    \end{minipage}
\end{table}

\noindent\textbf{Dual-branch Kinematic Attention.}
Removing the dynamic branch (Row 2) increases spatial errors but reduces Jitter, indicating a trade-off between spatial accuracy and temporal smoothness. Removing the structural branch (Row 3) increases MPJPE, MPJVE, and Jitter. These results suggest distinct contributions from the two branches: dynamic interactions primarily benefit spatial accuracy, whereas the structural branch provides modest improvements in positional and temporal metrics.

\noindent\textbf{Body Partitioning Strategies.} Partitioning the body into the torso and four limbs is common in full-body motion modeling, but its suitability for sparse-input reconstruction remains to be established. We therefore compare this design with four alternatives: no partitioning (Single body), upper/lower grouping, left/middle/right grouping, and five semantically shuffled groups (Shuffled). As shown in Table~\ref{tab:ablation_partition}, our partitioning achieves the lowest MPJRE, MPJPE, and MPJVE, supporting the effectiveness of anatomically meaningful grouping for sparse full-body reconstruction.

\begin{table}[t]
    \centering
    \caption{Evaluation of sensor scalability and masking strategies.}
    \label{tab:scalability}
    \small
    \setlength{\tabcolsep}{3pt}
    \begin{tabular}{lcccc}
        \toprule
        Configuration & MPJRE$\downarrow$ & MPJPE$\downarrow$ & MPJVE$\downarrow$ & Jitter$\downarrow$ \\
        \midrule
        \multicolumn{5}{c}{Decaying Masking} \\
        \midrule
        HMD           & 2.22 & 2.95 & 14.88 & 6.02 \\
        HMD+1 sensor  & 1.82 & 1.83 & 9.57  & 5.04 \\
        HMD+2 sensors & 1.55 & 1.18 & 5.86  & 3.63 \\
        HMD+3 sensors & \textbf{1.24} & \textbf{0.96} & \textbf{4.03} & \textbf{3.54} \\
        \midrule
        \multicolumn{5}{c}{Continuous Masking} \\
        \midrule
        HMD           & 2.26 & 2.97 & 15.37 & 6.24 \\
        HMD+1 sensor  & 2.09 & 2.49 & 13.22 & 5.79 \\
        HMD+2 sensors & 1.76 & 1.60 & 8.25  & 4.11 \\
        HMD+3 sensors & 1.60 & 1.51 & 7.58  & 4.15 \\
        \bottomrule
    \end{tabular}
\end{table}

\subsection{Scalability and Curriculum Strategy}
Our unified representation accommodates additional wearable sensors without changes to the network architecture. Following HMD-Poser~\cite{dai2024hmd}, we evaluate three extended tracking configurations: +1 (pelvis), +2 (both knees), and +3 (pelvis and both knees). For each configuration, we include the additional tracked joints in the target observed set and retain their observations as anchors during masking. As shown in Table~\ref{tab:scalability}, integrating additional sensors under our curriculum consistently improves reconstruction accuracy and motion smoothness, demonstrating the scalability of HolisticMotion across these tracking configurations.

We also compare our curriculum with a continuous masking baseline that uses randomly masked full-body inputs throughout training. A separate curriculum model is trained for each target configuration, whereas continuous masking uses a single shared model across all evaluated configurations. Our curriculum achieves lower errors across all four evaluation metrics in every configuration, highlighting a practical trade-off between configuration-specific reconstruction accuracy and the deployment flexibility of a single shared model.
\section{Conclusion and Limitations}
\label{sec:conclusion}
We presented HolisticMotion, a framework for reconstructing full-body motion from sparse tracking signals through holistic full-body modeling. By organizing observed and unobserved joints within a unified skeletal representation, our approach combines curriculum-based masked training, body partitioning, and Kinematic Attention to learn full-body motion dependencies. Experiments on AMASS and HPS demonstrate competitive reconstruction performance across the evaluated tracking protocols. The unified input layout also accommodates additional wearable sensors without architectural changes. The model achieves 84.5 FPS on an NVIDIA RTX 5060 Ti, exceeding the throughput required for 60 Hz tracking but falling short of 120 Hz. Future work will explore lightweight network architectures for deployment on resource-constrained devices.
\clearpage
\section*{LLM Usage Statement}
\label{sec:LLM}
We used OpenAI Codex and Google Gemini to assist with drafting, revising, and polishing manuscript text, and DeepSeek to assist with writing code for the research implementation. The authors reviewed and revised the assisted text and checked the assisted code, and take full responsibility for the final submission.
{
    \small
    \bibliographystyle{ieeenat_fullname}
    \bibliography{main}
}

% WARNING: do not forget to delete the supplementary pages from your submission 
\clearpage

\end{document}